\documentclass{article}
\usepackage{aikyam}						

\usepackage{booktabs}						
\usepackage{multirow}						
\usepackage{amsfonts}						
\usepackage{graphicx}						
\usepackage{duckuments}						
\usepackage{enumitem}

\usepackage[numbers,compress,sort]{natbib}	

\title[Towards Understanding the Robustness of Sparse Autoencoders]{Towards Understanding the Robustness of Sparse Autoencoders}

\author[Saiyed et al.]{%
Ahson Saiyed$^{*}$\\
University of Virginia \\\And
Sabrina Sadiekh\thanks{Equal Contribution. Corresponding authors: \href{mailto:as7ac@virginia.edu}{Ahson Saiyed} and \href{mailto:sadsobr7@gmail.com}{Sabrina Sadiekh}}\\
Independent Researcher \\\And
Chirag Agarwal\\
University of Virginia
}

\newcommand{\xhdr}[1]{\vspace{0em}\noindent{{\bf #1.}}}

\newcommand{\eg}{\textit{e.g., \xspace}}
\newcommand{\sae}{\textsc{Sae}\xspace}
\newcommand{\base}{\textsc{Base}\xspace}
\newcommand{\beast}{\textsc{Beast}\xspace}
\newcommand{\gcg}{\textsc{Gcg}\xspace}
\newcommand{\hide}[1]{}

\newcommand{\std}[1]{\scriptsize{$\pm$#1}}

\usepackage{colortbl}

\definecolor{Gray}{gray}{0.9}
\definecolor{LightCyan}{rgb}{0.88,1,1}
\definecolor{darkred}{rgb}{0.8,0.1,0.1}
\definecolor{darkyellow}{rgb}{0.95, 0.68, 0.22}
\definecolor{darkgreen}{rgb}{0.1,0.8,0.1}
\newcolumntype{a}{>{\columncolor{Gray}}c}
\newcolumntype{b}{>{\columncolor{white}}c}

\begin{document}

\maketitle
\begin{abstract}
    \looseness=-1
Large Language Models (LLMs) remain vulnerable to optimization-based jailbreak attacks that exploit internal gradient structure. While Sparse Autoencoders (\sae) are widely used for interpretability, their robustness implications remain underexplored. We present a study of integrating pretrained \sae into transformer residual streams at inference time, without modifying model weights or blocking gradients.  Across four model families (Gemma, LLaMA, Mistral, Qwen) and two strong white-box attacks (\gcg, \beast) plus three black-box benchmarks, \sae-augmented models achieve up to a 5$\times$ reduction in jailbreak success rate relative to the undefended baseline and reduce cross-model attack transferability. Parametric ablations reveal (i) a monotonic dose–response relationship between $L_0$ sparsity and attack success rate, and (ii) a layer-dependent defense–utility tradeoff, where intermediate layers balance robustness and clean performance.  These findings are consistent with a representational bottleneck hypothesis: sparse projection reshapes the optimization geometry exploited by jailbreak attacks. Code and data are available on \href{https://github.com/AikyamLab/sparse-jailbreak}{GitHub} and \href{https://huggingface.co/datasets/Aikyam-Lab/Sparse-JailBreak-GCG-Suffixes}{HuggingFace}.    
\end{abstract}

\section{Introduction}
\label{sec:intro}

Large Language Models (LLMs) are increasingly deployed in safety-critical settings, yet they remain vulnerable to \emph{jailbreaks}: carefully constructed inputs that elicit unsafe or policy-violating behavior~\cite{yi2024jailbreakattacksdefenseslarge}. Many state-of-the-art jailbreak attacks explicitly exploit gradient-aligned directions in internal activations, suggesting that adversarial success is tightly coupled to the geometry of intermediate representations.

Recent advances in mechanistic interpretability have revealed that transformer residual streams are not amorphous vector spaces but exhibit structured latent features~\cite{latentProbingContextual, skean2024doesrepresentationmatterexploring, yugeswardeenoo2025interpretinglatentstructureoperator}. Sparse Autoencoders (\sae) provide a practical method for decomposing dense activations into sparse, feature-aligned components that often correspond to interpretable concepts~\cite{cunningham2023sparse, li2025geometryconceptssparseautoencoder}. By reparameterizing activations in a sparse basis, \sae impose a structured bottleneck on the residual stream.
However, the robustness implications of \sae-based representations remain unclear. Prior work has shown that \sae features can be sensitive to small perturbations~\cite{li2025interpretabilityillusionssparseautoencoders}, and that reconstruction errors may induce behavioral shifts when reinjected into the model~\cite{gurnee2024pathological}. These findings raise concerns about using \sae for monitoring or control, but they do not address a complementary question: \textit{how does inserting sparse feature projections into LLM residual streams affect adversarial robustness under adaptive jailbreak attacks?}

\noindent\looseness=-1\textbf{Present Work.} We conduct a controlled empirical study of inserting pretrained \sae into transformer layers at inference time. The intervention does not modify model weights, does not block gradients, and preserves end-to-end differentiability, allowing us to isolate the representational effect of sparse projection on adversarial optimization dynamics.

\looseness=-1 We study \sae insertion across three open-source model families (Gemma, LLaMA, and Mistral). To further examine how robustness depends on insertion depth, we additionally analyze Qwen2.5 using a multi-layer \sae suite that enables controlled layer-placement experiments. Robustness is measured under strong white-box jailbreak attacks (\gcg and \beast) as well as black-box jailbreak benchmarks. Across evaluation settings, \sae-augmented models reduce jailbreak success rates by up to $\mathbf{5\times}$ relative to the undefended baseline.

Beyond aggregate performance, we investigate how robustness varies with \sae configuration through sparsity and layer-placement ablations, and analyze cross-model transferability of adversarial suffixes. Our results show that \sae routing systematically reduces attack transferability and alters the geometry of adversarial optimization, suggesting that structured sparse projections can act as lightweight representation-level defenses against optimization-based jailbreak attacks.

\section{Related works}
\label{sec:related}
\looseness=-1

Our work connects two lines of research: adversarial jailbreak attacks on LLMs and \sae for mechanistic interpretability.

\xhdr{Adversarial Jailbreak Attacks}
Early adversarial NLP work focused on discrete perturbations such as token substitutions~\cite{ebrahimi2018hotflip} and paraphrasing~\cite{alzantot2018generating}. More recent jailbreak methods directly target alignment objectives. 
\citet{zou2023universal} introduced Greedy Coordinate Gradient (GCG), a gradient-based adversarial suffix optimizer with strong cross-model transfer. BEAST improves attack efficiency via beam-search heuristics~\cite{sadasivan2024fast}. Extensions include prompt injection~\cite{perez2022ignore}, indirect exploits~\cite{greshake2023not}, and automated multi-turn attacks~\cite{reddy2025autoadvautomatedadversarialprompting}. These methods exploit the smooth internal optimization landscape of LLMs, enabling gradient-aligned suffix construction under white-box access.

\xhdr{Sparse Autoencoders for Interpretability}
Mechanistic interpretability aims to decompose dense transformer activations into structured, interpretable components. \sae reconstruct activations as sparse linear combinations of latent features~\cite{bricken2023monosemanticity,cunningham2023sparse}, and have been scaled to large models with high reconstruction fidelity~\cite{gao2024scaling,templeton2024scaling}. \sae enable feature discovery, attribution, and feature-level interventions (\eg steering or concept erasure)~\cite{marks2025sparsefeaturecircuitsdiscovering,obrien2025steeringlanguagemodelrefusal,cheng2026toward}. Recent work has also identified limitations of \sae representations. 
\citet{li2025interpretabilityillusionssparseautoencoders} show that \sae latents can be manipulated by small input perturbations, challenging their stability as monitoring tools. 
\citet{gurnee2024pathological,engels2025darkmatter} demonstrate that reconstruction errors may induce structured semantic shifts. 
These findings raise open questions about the robustness properties of \sae-based representations when embedded into model pipelines.

\looseness=-1\xhdr{LLM Defense} A variety of defenses have been proposed against jailbreak attacks~\cite{yi2024jailbreakattacksdefenseslarge}. Randomized smoothing methods such as Erase-and-Check~\cite{kumar2025certifyingllmsafetyadversarial}, SmoothLLM~\cite{robey2024smoothllmdefendinglargelanguage}, and Semantic Smoothing~\cite{ji2024defendinglargelanguagemodels} apply stochastic perturbations or auxiliary paraphrasing models. Noise-based approaches such as Smoothed Embeddings~\cite{hase2025smoothedembeddingsrobustlanguage} trade off attack success rate (ASR) and utility via input-level noise injection. Other methods intervene at decoding time~\cite{zhao2026defendinglargelanguagemodels} or rely on finetuning~\citep{han2024medsafetybench} and adversarial training~\cite{fu2026shortlengthadversarialtraininghelps}. 

\xhdr{\textit{Research Gap}} These approaches primarily evaluate aggregate ASR and utility trade-offs. In contrast, we study how inserting a sparse encode–decode bottleneck into the residual stream alters adversarial optimization dynamics and cross-model transferability. A detailed comparison with representative smoothing-, noise-, and decoding-based defenses is provided in Appendix~\ref{app:comparing_with_other_defenses}.
\section{Method}
\label{sec:method}

We study an activation-space intervention against adversarial suffix attacks by inserting pretrained Sparse Autoencoders (\sae) into intermediate transformer layers. The \sae acts as a sparse encode--decode operator applied to the residual stream during inference. Model weights remain unchanged and gradients are fully preserved, allowing white-box attackers to optimize directly through the \sae transformation. Our evaluation spans four model families (Gemma-2, LLaMA-3, Mistral, and Qwen2.5) and scales from 2B to 70B parameters. Robustness is tested under strong white-box jailbreak attacks (GCG and \beast) and black-box jailbreak benchmarks.

\subsection{\sae-Augmented Language Models}
\label{sec:models_setup}

Let $\mathbf{h}_\ell \in \mathbb{R}^{d}$ denote the residual stream activation at transformer layer $\ell$. We insert a pretrained Sparse Autoencoder (\sae) that applies a sparse encode--decode transformation to the activation:
\[
\hat{\mathbf{h}}_\ell = \mathcal{D}(\mathcal{E}(\mathbf{h}_\ell)),
\]
where the encoder and decoder are defined as
$ \mathbf{z} = \mathcal{E}(\mathbf{h}) = \mathrm{ReLU}(W_{\text{enc}}\mathbf{h} + \mathbf{b}_{\text{enc}}),$ 
$\hat{\mathbf{h}} = \mathcal{D}(\mathbf{z}) = W_{\text{dec}}\mathbf{z} + \mathbf{b}_{\text{dec}}.$

The latent representation $\mathbf{z} \in \mathbb{R}^{d_{\text{hidden}}}$ is sparse, with $d_{\text{hidden}} \gg d$ (typically a $16\times$ expansion). \sae are trained offline to minimize reconstruction error with an $\ell_1$ sparsity penalty, yielding a sparse feature basis that approximately reconstructs the original residual activation.

Let $f_\theta$ denote the base language model and let $\Phi_\ell(\cdot)$ denote the \sae routing inserted at layer $\ell$. The resulting \sae-augmented model is defined as
\begin{equation}
    \tilde{f}_\theta(x)
    =
    f_\theta^{>\ell}
    \big(
    \Phi_\ell(
    f_\theta^{\le \ell}(x)
    )
    \big),
\end{equation}
where $f_\theta^{\le \ell}$ denotes the transformer prefix up to layer $\ell$ and $f_\theta^{>\ell}$ the remaining layers. The routing operator $\Phi_\ell(\mathbf{h}_\ell) = \mathcal{D}(\mathcal{E}(\mathbf{h}_\ell))$ replaces the residual activation at that layer during inference: $\mathbf{h}_\ell \rightarrow \hat{\mathbf{h}}_\ell.$ All subsequent transformer blocks therefore operate on the reconstructed activation $\hat{\mathbf{h}}_\ell$.

Importantly, the \sae layer remains fully differentiable. Under white-box attacks, adversarial suffix tokens $x_{\text{adv}}$ are optimized to minimize the attack objective
\begin{equation}
    \min_{x_{\text{adv}}}
    \mathcal{L}\big(
    \tilde{f}_\theta(x_{\text{prompt}} \oplus x_{\text{adv}})
    \big).
\end{equation}

Because the routing function $\Phi_\ell$ is differentiable, gradients propagate through the \sae reconstruction during optimization. For gradient-based attacks such as GCG, suffix tokens are updated using gradients that pass through the \sae layer at each step:
\begin{equation}
    \nabla_{x_{\text{adv}}}
    \mathcal{L}
    =
    \frac{\partial \mathcal{L}}{\partial \hat{\mathbf h}_\ell}
    \frac{\partial \hat{\mathbf h}_\ell}{\partial \mathbf h_\ell}
    \frac{\partial \mathbf h_\ell}{\partial x_{\text{adv}}}.
\end{equation}

Thus adversarial optimization proceeds through the \sae-modified representation space rather than the original residual stream. In contrast, gradient-free attacks such as \beast construct adversarial suffixes via beam search without accessing gradients; candidate suffixes are therefore evaluated directly on the \sae-augmented model $\tilde f_\theta$. While the optimization procedures differ, both attacks operate over the same modified representation induced by the \sae routing.

\subsection{Model Suite}
\label{sec:models_families}
We evaluate \sae insertion across multiple open-source LLMs: Gemma-2 (2B, 9B, 27B), LLaMA-3 (8B, 70B), Mistral-7B, and Qwen2.5-7B. For Gemma, LLaMA, and Mistral we insert a single \sae into an intermediate transformer layer (layer mappings in Appendix~\ref{tab:sae2}). To analyze the effect of insertion depth, we additionally evaluate Qwen2.5 using a multi-layer \sae suite that enables controlled layer-placement experiments.

\subsection{Threat Model}
\label{sec:attacks}
We consider adversaries that append adversarial suffixes to prompts in order to induce harmful model outputs.

\xhdr{White-box attacks} We evaluate two white-box jailbreak methods. Greedy Coordinate Gradient (\gcg)~\cite{zou2023universal} iteratively optimizes suffix tokens using gradients from a harmful objective. In our setting, gradients are backpropagated through the \sae layer, meaning the attacker directly optimizes through the sparse representation. \beast~\cite{sadasivan2024fast} is a beam-search-based jailbreak method that leverages model logits and internal signals. This provides a complementary non-gradient attack that still assumes white-box access.

\paragraph{Black-box attacks.}
We additionally evaluate robustness under query-only access using 1,500 jailbreak prompts drawn from Salad-Data~\cite{li2024salad}, Prompt Injections Benchmark, and SafeEval~\cite{manczak2024primeguard}.

\subsection{Implementation Details}

All models are run in PyTorch with bfloat16 precision on NVIDIA GPUs. Special tokens are excluded from activation analysis. We evaluate 24 HarmBench batches and report aggregated ASR and mechanistic metrics.
\section{Experiments}

We conduct a systematic empirical evaluation of \sae insertion as an inference-time intervention against adversarial suffix attacks. Our experiments address four research questions:

\looseness=-1\noindent\textbf{RQ1 (Effectiveness).} Does inserting a pretrained \sae reduce attack success rates under adaptive white-box and black-box jailbreak attacks?

\looseness=-1\noindent\textbf{RQ2 (Transferability).} How does \sae insertion affect the cross-model and cross-configuration transferability of adversarial suffixes?

\looseness=-1\noindent\textbf{RQ3 (Parametric Dependence).} How do robustness outcomes depend on \sae configuration, including sparsity level ($L_0$) and insertion depth?

\looseness=-1\noindent\textbf{RQ4 (Optimization Dynamics).} How does \sae insertion alter adversarial optimization dynamics and sparse feature usage?

\subsection{Experimental Setup}

\xhdr{Model suite}
We evaluate \sae insertion across six open-source models spanning three families and a wide range of scales: Gemma-2 (2B, 9B, 27B), LLaMA-3 (8B, 70B), and Mistral-7B. For each model, a pretrained \sae is inserted into a single intermediate transformer layer following the procedure described in Section~\ref{sec:method}. Layer mappings are reported in Appendix Table~\ref{tab:sae2}.

To study the effect of insertion depth, we additionally perform controlled layer-placement experiments on Gemma-2-9B and Qwen2.5 using a multi-layer \sae suite.

\looseness=-1\xhdr{Attack protocol} We evaluate robustness under both gradient-based and non-gradient white-box jailbreak attacks. For \gcg, adversarial suffixes are optimized for 500 gradient steps with a fixed suffix length of 20 tokens. For \beast, we use beam-search parameters $k_1{=}k_2{=}15$ and search depth $L{=}20$, following the original attack configuration.

Each prompt is evaluated under three configurations: 
\begin{itemize}[itemsep=0pt, parsep=0pt, topsep=2pt]
\item \textsc{Prompt}: clean harmful prompt without an adversarial suffix,
\item \base: suffix optimized against the baseline model,
\item \sae: suffix optimized against the \sae-augmented model.
\end{itemize}

This protocol allows us to evaluate baseline vulnerability, adaptive attack performance, and cross-configuration transferability. In addition to white-box attacks, we evaluate robustness under black-box access using three jailbreak datasets described in Section~\ref{sec:attacks}. In this setting, attackers interact with the model through prompt queries only.

\begin{figure*}[t]
\centering
\includegraphics[width=\textwidth]{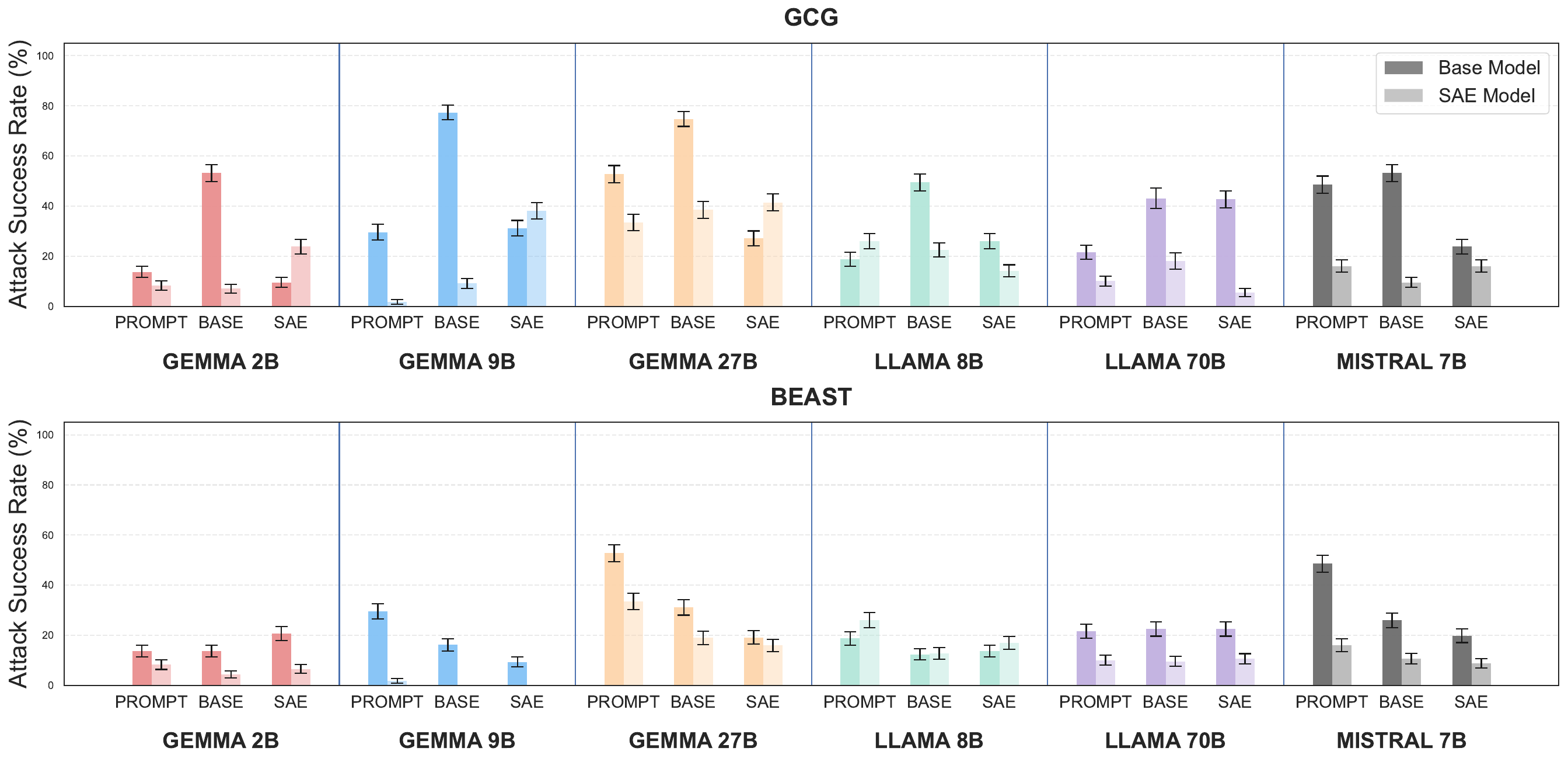}
\caption{Attack success rate (ASR) for \gcg and \beast attacks on HarmBench across baseline (dark) and \sae-augmented (light) models. Results are shown for six models (Gemma-2 2B/9B/27B, LLaMA-3 8B/70B, Mistral-7B) under three configurations: \textsc{Prompt}, \base, and \sae. Under adaptive attacks, \sae-augmented models exhibit substantially lower ASR than their baseline counterparts. Under \base transfer evaluation, median ASR decreases from 55.0\% to 19.05\% for \gcg and from 19.35\% to 9.7\% for \beast. Error bars denote standard error over HarmBench prompts ($n \approx 218$ per model--condition).}
\label{fig:asr_paired}
\end{figure*}

\subsection{Evaluation Metrics}

Our primary metric is attack success rate (ASR), defined as the fraction of prompts that produce harmful outputs under evaluation. Harmful responses are detected using three independent evaluators: WildGuard, the HarmBench classifier, and a refusal-based heuristic (Appendix~\ref{tab:refusal_patterns}). Beyond aggregate ASR, we analyze cross-model transferability of adversarial suffixes and examine representation-level changes induced by \sae insertion.

\subsection{Mechanistic Analyses}

To study how \sae routing alters adversarial representations and optimization dynamics, we perform two additional analyses.

\paragraph{Sparse feature analysis.}
We analyze \sae latent activations $\mathbf{z}$ induced by adversarial suffixes. For each prompt we extract the top-$k$ activated features and measure similarity using the Jaccard index:

\[
J(A,B) = \frac{|A \cap B|}{|A \cup B|}.
\]

\paragraph{Gradient spectral analysis.}
To characterize adversarial optimization dynamics under \gcg, we compute the singular value decomposition of the gradient matrix $G \in \mathbb{R}^{L \times d}$ for suffix length $L$. We measure effective rank, spectral gap, and inter-step cosine similarity to quantify changes in gradient concentration.
\section{Results}
\label{sec:results}

\paragraph{RQ1: Does \sae insertion reduce attack success rates under adaptive white-box attacks?}

\xhdr{White-box (\gcg)} We first evaluate the effectiveness of \sae insertion under adaptive white-box attacks, where adversarial suffixes are optimized directly against the evaluation model. As summarized in Table~\ref{tab:asr_median_tests} and illustrated in Fig.~\ref{fig:asr_paired}, \sae-augmented models exhibit substantially lower attack success rates than their baseline counterparts. Across the six evaluated models, the median ASR decreases from 55.0\% for baseline models to 19.05\% for \sae-augmented models. This reduction is statistically significant under a two-sided Mann--Whitney U test ($p=0.0087$). Paired ASR scores for both attacks are reported in Appendix~\ref{app:asr}.

\looseness=-1\xhdr{White-box (\beast)} A similar trend is observed under the non-gradient \beast attack. Median ASR decreases from 19.35\% for baseline models to 9.7\% for \sae-augmented models (Table~\ref{tab:asr_median_tests}, Fig.~\ref{fig:asr_paired}). This reduction is statistically significant under the Mann--Whitney U test ($p=0.0411$), indicating that \sae insertion improves robustness even for attacks that do not rely on gradient-based optimization.

\xhdr{Black-box benchmarks}
We additionally evaluate \sae robustness under black-box jailbreak prompts using three external datasets (Appendix Table~\ref{tab:blackbox_asr}). Across benchmarks, \sae-augmented models consistently exhibit lower ASR than their baseline counterparts, although the magnitude of improvement varies depending on the evaluator and prompt distribution.

\begin{table}[t]
\centering
\small
\caption{Median ASR (\%) for adversarial suffixes optimized against the evaluation model and evaluated on baseline versus \sae-augmented models ($N=6$). Statistical significance is assessed using a two-sided Mann--Whitney U test. An asterisk (*) indicates statistical significance at $p < 0.05$.}
\begin{tabular}{lccc}
\toprule
Attack & \base & \sae & MWU ($p$) \\
\midrule
\gcg   & 55.0 & 19.05 & 0.0087* \\
\beast & 19.35 & 9.7  & 0.0411* \\
\bottomrule
\end{tabular}
\label{tab:asr_median_tests}
\end{table}

\xhdr{RQ2 (Transferability): \sae insertion reduces cross-model and cross-configuration transfer} \looseness=-1 We analyze transferability of adversarial suffixes under four source--target configurations:
\base$\rightarrow$\base,
\sae$\rightarrow$\sae,
\base$\rightarrow$\sae, and
\sae$\rightarrow$\base.
We distinguish \emph{same-configuration cross-model transfer} (\base$\rightarrow$\base and \sae$\rightarrow$\sae) from \emph{cross-configuration transfer} (\base$\rightarrow$\sae and \sae$\rightarrow$\base). For \base$\rightarrow$\base and \sae$\rightarrow$\sae, we exclude same-model attacks (the main diagonal) and therefore evaluate only off-diagonal pairs; with six models this yields $6\times5=30$ directed source--target pairs.
For \base$\rightarrow$\sae and \sae$\rightarrow$\base, each \base/\sae variant of a model is treated as a distinct source/target, so we evaluate all off-diagonal cross-configuration pairs, yielding $6\times6=36$ directed pairs per direction.\footnote{All transfer analyses exclude same-model attacks; see Appendix for full matrices.}

\looseness=-1\xhdr{\gcg} Under \gcg, \sae insertion reduces cross-model transfer within the same configuration: \sae$\rightarrow$\sae exhibits a lower median ASR than \base$\rightarrow$\base (2.65\% vs.\ 8.75\%; Table~\ref{tab:transfer_gcg_beast}), and the confidence intervals also shift downward (\base$\rightarrow$\base: [6.25, 13.75] vs.\ \sae$\rightarrow$\sae: [1.25, 8.20]). Cross-configuration transfer further reveals a pronounced asymmetry: suffixes optimized on \base models transfer less effectively to \sae targets than \sae-optimized suffixes transfer to base targets (\base$\rightarrow$\sae median 7.50\%, CI [2.50, 10.00] vs.\ \sae$\rightarrow$\base median 12.65\%, CI [11.00, 17.90]). Notably, the \base$\rightarrow$\sae interval lies substantially below \sae$\rightarrow$\base, indicating that \sae insertion weakens transferred attacks from baseline models while \sae-optimized suffixes remain comparatively more transferable to baseline targets.

\looseness=-1\xhdr{\beast} A similar pattern holds for the non-gradient \beast attack. Within-class cross-model transfer is lower for \sae models (\sae$\rightarrow$\sae median 5.35\%) than for baseline models (\base$\rightarrow$\base median 15.25\%), with non-overlapping shifts in the corresponding confidence intervals (\base$\rightarrow$\base: [11.85, 19.30] vs.\ \sae$\rightarrow$\sae: [3.15, 9.15]). Cross-configuration transfer is again asymmetric: \base$\rightarrow$\sae achieves a median ASR of 8.20\% (CI [5.20, 11.00]) whereas \sae$\rightarrow$\base achieves 11.90\% (CI [9.40, 16.95]). Compared to \gcg, the asymmetry is weaker but still consistent in direction, suggesting that \sae insertion reduces transferability from baseline to defended targets even when the attacker does not rely on dense gradients.

\looseness=-1 Overall, these results show that \sae insertion reduces cross-model transfer (\base$\rightarrow$\base to \sae$\rightarrow$\sae shift) and induces an asymmetric cross-configuration transfer pattern (\base$\rightarrow$\sae $<$ \sae$\rightarrow$\base), consistent with the view that \sae routing changes the geometry of transferable adversarial solutions rather than acting as output filtering.

\begin{figure*}[t]
    \centering
    \includegraphics[width=\textwidth]{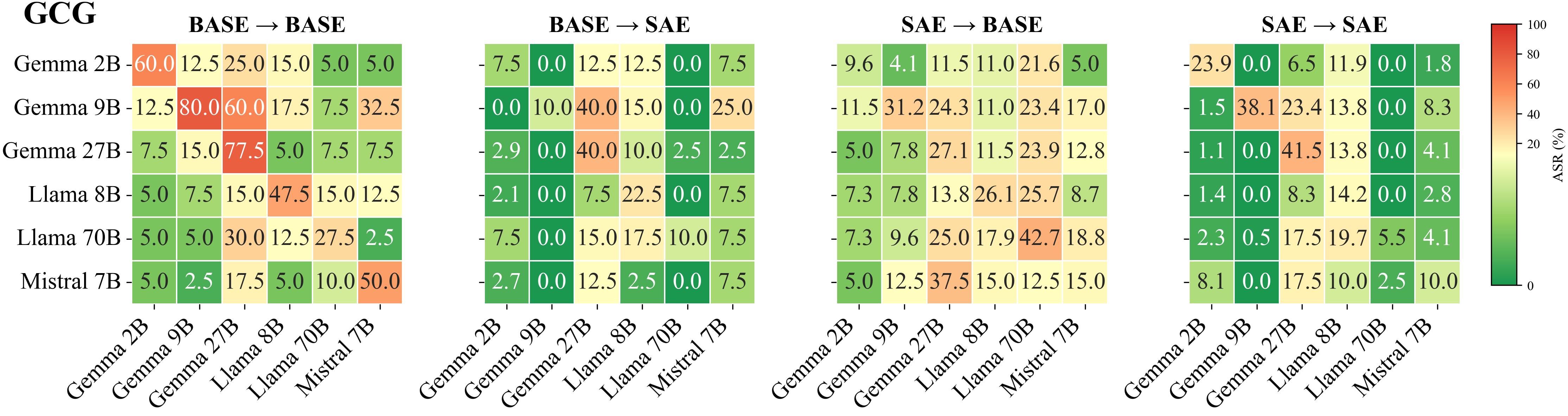}\\[0.3em]
    \includegraphics[width=\textwidth]{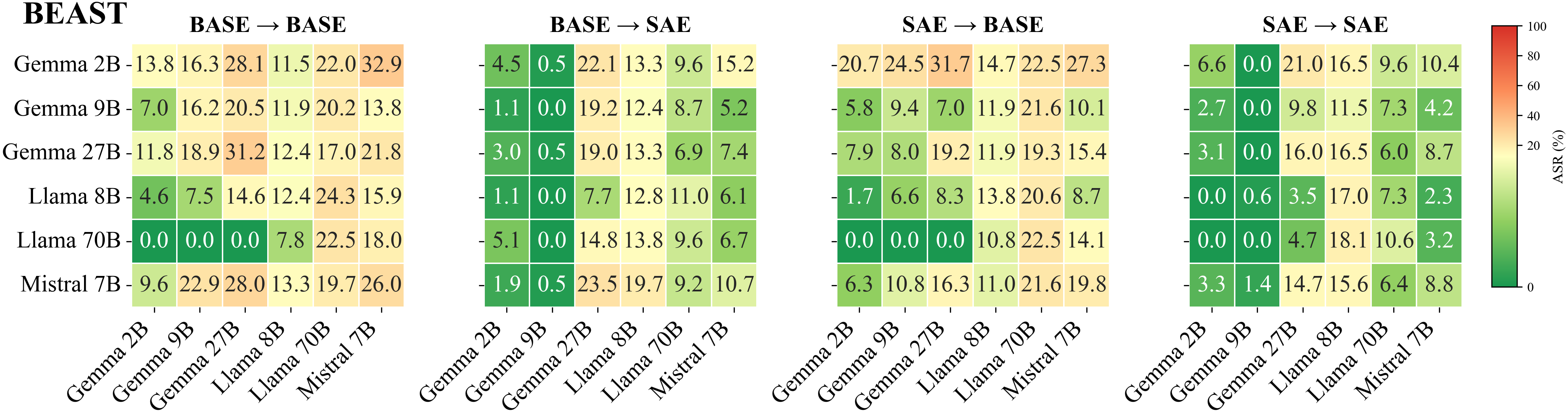}
    \caption{Attack Success Rate (ASR) transfer matrices for \gcg (top) and \beast (bottom) attacks across evaluation models. Rows correspond to the suffix source model used to generate adversarial suffixes, while columns denote the evaluation (target) models. Values are normalized to the range $[0,100]$, where higher values indicate a higher attack success rate.}
    \label{fig:ASR_matrix_combined}
\end{figure*}



\begin{table}[t]
\centering
\small
\caption{Cross-model transfer of median ASR (\%) for \gcg and \beast. \base$\rightarrow$\base and \sae$\rightarrow$\sae exclude same-model attacks and therefore contain $6\times5=30$ directed off-diagonal pairs. \base$\rightarrow$\sae and \sae$\rightarrow$\base evaluate all directed cross-configuration pairs ($6\times6=36$). Confidence intervals are 95\% bootstrap intervals over model pairs.}
\begin{tabular}{llccc}
\toprule
\textbf{Attack} & \textbf{Transfer} & \textbf{ASR} & \textbf{95\% CI} & \textbf{N} \\
\midrule
\multirow{4}{*}{GCG}
 & \base$\rightarrow$\base & 8.75  & [6.25, 13.75] & 30 \\
 & \sae$\rightarrow$\sae   & 2.65  & [1.25, 8.20]  & 30 \\
 & \base$\rightarrow$\sae  & 7.50  & [2.50, 10.00] & 36 \\
 & \sae$\rightarrow$\base  & 12.65 & [11.00, 17.90] & 36 \\
\midrule
\multirow{4}{*}{\beast}
 & \base$\rightarrow$\base & 15.25 & [11.85, 19.30] & 30 \\
 & \sae$\rightarrow$\sae   & 5.35  & [3.15, 9.15]   & 30 \\
 & \base$\rightarrow$\sae  & 8.20  & [5.20, 11.00]  & 36 \\
 & \sae$\rightarrow$\base  & 11.90 & [9.40, 16.95]  & 36 \\
\bottomrule
\end{tabular}
\label{tab:transfer_gcg_beast}
\end{table}

\looseness=-1\xhdr{RQ3 (Parametric Dependence): Robustness varies with sparsity and insertion depth} We perform ablations over two key parameters: sparsity level ($L_0$) and insertion depth. We focus on model families for which pretrained open-source \sae suites are available (Gemma-2, LLaMA-3.1), enabling controlled interventions without retraining. Because our primary robustness signal arises from adversarial transfer, we focus on transfer settings (\base$\rightarrow$\sae and \sae$\rightarrow$\base) rather than within-model attacks.

\xhdr{Sparsity ablation} We vary \sae sparsity while keeping the insertion layer fixed and evaluate its effect on adversarial transfer. Across both Gemma-2 9B and LLaMA-3.1-8B-Instruct, we observe a consistent monotonic relationship between sparsity and robustness.

For Gemma-2 9B, increasing $L_0$ (lower sparsity) leads to a substantial degradation in robustness: \base$\rightarrow$\sae ASR increases from 0.9\% at $L_0=11$ to 21.2\% at $L_0=310$ (Table~\ref{tab:sparsity}). A similar trend holds for reverse transfer, where \sae$\rightarrow$\base ASR increases from 26.0\% to 43.8\%. We observe the same pattern on LLaMA-3.1-8B-Instruct when varying the number of active features $k$ (a proxy for effective sparsity). Under \base$\rightarrow$\sae, ASR remains low at 0.5\% for $k=32$ and $k=64$, but increases to 2.3\% for $k=128$ and $k=256$ (Table~\ref{tab:llama_sparsity}). Reverse transfer again follows the same trend, with \sae$\rightarrow$\base ASR increasing from 35.0\% to 48.2\%.

Together, these results reveal a consistent dose–response relationship: stronger sparse compression more effectively disrupts transferable adversarial representations, while relaxing sparsity restores transferability across models.

\begin{table}[t]
\centering
\small
\caption{\looseness=-1 Effect of \sae sparsity ($L_0$) on transfer ASR (\%) for Gemma-2 9B (Layer 20, Width 16K). Each cell reports $\approx$218 prompts. The monotonic increase in \base$\rightarrow$\sae ASR with larger $L_0$ indicates that stronger sparse compression more effectively disrupts transferable adversarial representations.}
\begin{tabular}{rcc}
\toprule
\textbf{$L_0$} & \base$\rightarrow$\sae & \sae$\rightarrow$\base \\
\midrule
11  & \textbf{0.9}  & 26.0 \\
36  & 3.2  & 18.1 \\
68  & 14.4 & 28.8 \\
138 & 15.6 & 35.8 \\
310 & 21.2 & \textbf{43.8} \\
\bottomrule
\end{tabular}
\label{tab:sparsity}
\end{table}

\begin{table}[t]
\centering
\small
\caption{Effect of \sae sparsity (controlled via top-$k$ active features) on transfer ASR (\%) for LLaMA-3.1-8B-Instruct (Layer 19, Width 131K). Increasing $k$ (lower sparsity) degrades robustness, consistent with Gemma results.}
\begin{tabular}{rccc}
\toprule
\textbf{$k$} & \base$\rightarrow$\sae & \sae$\rightarrow$\base \\
\midrule
32 & \textbf{0.5} & 35.0 \\
64  & \textbf{0.5} & 38.2 \\
128 & 2.3 & 43.1 \\
256 & 2.3 & \textbf{48.2} \\
\bottomrule
\end{tabular}
\label{tab:llama_sparsity}
\end{table}

\xhdr{Layer placement ablation} We next vary insertion depth while keeping sparsity fixed, and observe a consistent dependence of robustness on representational depth across both Gemma-2 9B and LLaMA-3.1-8B-Instruct (with similar trends observed on Qwen2.5; see Appendix~\ref{app:qwen}).

Early-to-mid layers yield the strongest suppression of transfer attacks. For Gemma-2 9B, \base$\rightarrow$\sae ASR drops to 0.9\% at Layer 5 (Table~\ref{tab:layer_ablation}), while for LLaMA-3.1-8B-Instruct the best-performing layer is Layer 7 with 0.0\% ASR (Table~\ref{tab:llama_layer}). This indicates that intervening before adversarial features are fully formed can significantly disrupt transfer.

\looseness=-1 However, very early insertion can introduce behavioral side effects. In Gemma, early layers increase No Suffix ASR (12.2\% at Layer 5), while in LLaMA late layers exhibit even stronger degradation, with No Suffix ASR rising to 23.9\% at Layer 27. Intermediate layers provide a more favorable tradeoff. For example, in Gemma, Layer 20 maintains low transfer ASR (13.9\%) while preserving clean behavior, and in LLaMA, Layers 11–19 achieve similarly low transfer with minimal degradation.

In contrast, late-layer insertion substantially weakens the defense. In Gemma, \base$\rightarrow$\sae ASR increases to 51.4\% at Layer 35, while in LLaMA it rises to 7.3\% at Layer 23 and remains elevated for deeper layers. Reverse transfer (\sae$\rightarrow$\base) also increases with depth in both models, indicating that adversarial signals become more stable in later representations.

Overall, these results reveal a consistent layer-dependent defense–utility tradeoff: early layers maximize robustness but may perturb model behavior, intermediate layers offer the best balance, and late layers are largely ineffective.

\begin{table}[t]
\centering
\small
\caption{Effect of \sae layer placement on transfer ASR (\%) for Gemma-2 9B (Width 16K). \base denotes suffixes optimized on the base model and \sae suffixes optimized on the \sae model. \base$\rightarrow$\base baseline ASR = 77.3\%. Each cell reports $\approx$218 prompts.}
\begin{tabular}{lcc}
\toprule
\textbf{Layer} & \base$\rightarrow$\sae & \sae$\rightarrow$\base \\
\midrule
5  & \textbf{0.9}  & 13.4 \\
10 & 3.7  & 15.3 \\
20 & 13.9 & 30.5 \\
30 & 8.1  & 40.6 \\
35 & 51.4 & \textbf{53.2} \\
\bottomrule
\end{tabular}
\label{tab:layer_ablation}
\end{table}

\begin{table}[h]
\centering
\small
\caption{Effect of \sae layer placement on ASR (\%) for LLaMA-3.1-8B-Instruct (Width 131K). Early layers provide strongest robustness, while deeper layers degrade both robustness and clean performance.}
\begin{tabular}{ccccc}
\toprule
{Layer} & \base$\rightarrow$\sae & \sae$\rightarrow$\base \\
\midrule
3   & 0.5 & 19.7  \\
7   & \textbf{0.0} & 34.4 \\
11  & 0.9 & \textbf{45.9}  \\
15  & 0.5 & 39.2  \\
19  & 0.5 & 35.9  \\
23  & 7.3 & 38.1  \\
27  & 6.0 & 28.6  \\
\bottomrule
\end{tabular}
\label{tab:llama_layer}
\end{table}

\looseness=-1\xhdr{RQ4 (Optimization Dynamics): \sae insertion alters adversarial feature usage and gradient dynamics} To understand why \sae insertion reduces adversarial robustness and transferability, we analyze how it affects both the representation space used by attacks and the optimization dynamics of adversarial suffix generation.

\begin{figure}[t]
    \centering
    \includegraphics[width=\textwidth]{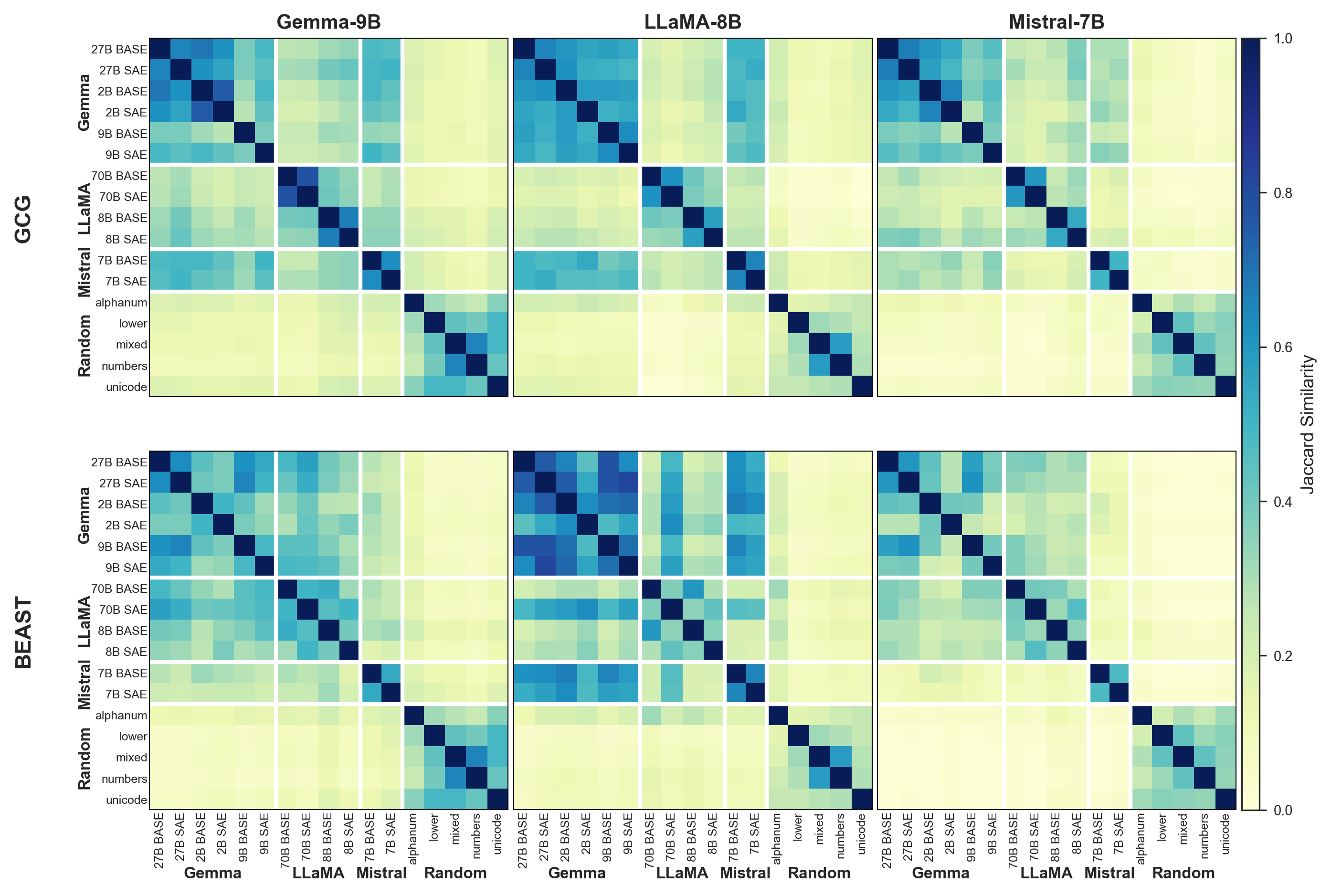}
    \caption{Adversarial suffixes from different source models exhibit high pairwise Jaccard similarity (0.36$\pm$0.16), while similarity to random baselines remains low (0.09$\pm$0.05). This 4$\times$ difference reveals that attacks converge on a shared sparse feature subspace, explaining their cross-model transferability.}
    \label{fig:gcg_jaccard}
\end{figure}

\begin{figure}[t]
\centering
\includegraphics[width=0.81\textwidth]{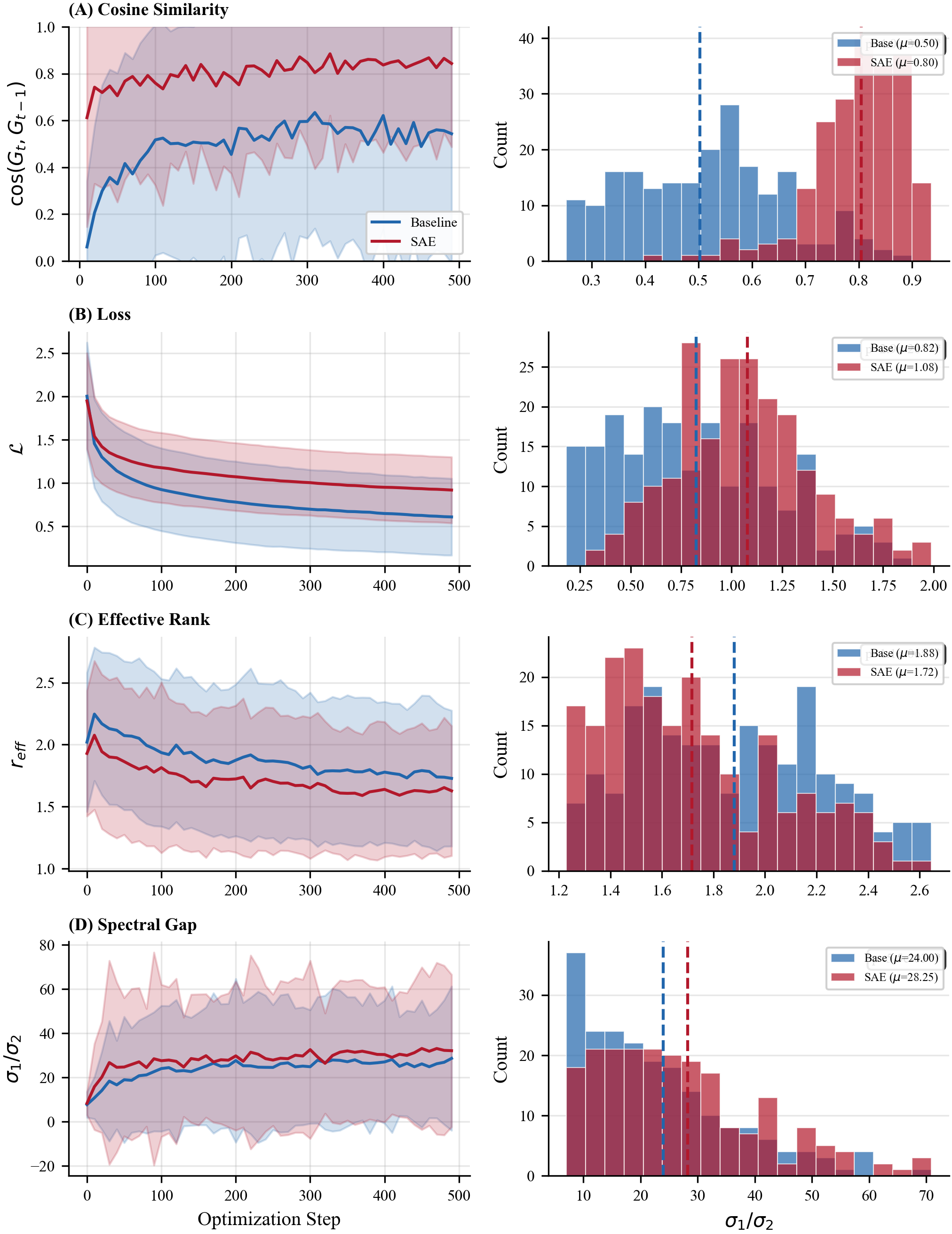}
t\caption{Spectral analysis of \gcg gradients on LLaMA-8B ($n$=218 paired runs). Left panels show metric trajectories over 500 optimization steps (mean ± std); right panels show distributions of trajectory means. \textbf{(A)} Cosine similarity: \sae gradients are 60\% more correlated between successive steps, indicating optimization stagnation. \textbf{(B)} Loss: \sae converges to 31\% higher loss. \textbf{(C)} Effective rank: \sae gradients occupy 9\% fewer dimensions. \textbf{(D)} Spectral gap: \sae's dominant singular value is 18\% stronger relative to the second. All differences significant (Wilcoxon $p < 0.01$).}
\label{fig:spectral}
\end{figure}
\looseness=-1\xhdr{Sparse feature structure of adversarial suffixes} We first examine which \sae features are activated by adversarial suffixes. For each suffix we extract the top-$k$ activated \sae features and measure overlap using the Jaccard index. 
As shown in Fig.~\ref{fig:gcg_jaccard}, adversarial suffixes exhibit substantially higher feature overlap with each other ($0.36 \pm 0.16$) than with random text baselines ($0.09 \pm 0.05$), a $4\times$ difference ($p<0.001$). This indicates that adversarial optimization does not produce arbitrary token perturbations, but instead converges on a shared sparse feature subspace. Within model families, feature overlap is consistently higher than across families, suggesting that different model architectures learn partially aligned but not identical adversarial feature representations. These structured feature patterns provide a mechanistic explanation for the cross-model transferability observed in RQ2.

\xhdr{Gradient dynamics during adversarial optimization} We next analyze gradient trajectories during \gcg optimization. For each attack run we compute spectral statistics of the token-level gradient matrix across optimization steps (Table~\ref{tab:spectral}; Fig.~\ref{fig:spectral}).

\begin{table}[h]
\centering
\small
\setlength{\tabcolsep}{4pt}
\caption{\looseness=-1 Spectral analysis of \gcg gradients on LLaMA-8B ($n$=218 paired attack runs, 10,900 gradient samples per condition, Wilcoxon signed-rank test). \sae integration significantly increases gradient similarity between steps (+60\%), while modestly reducing effective rank (-8.7\%).}
\begin{tabular}{lcccc}
\toprule
\textbf{Metric} & \base & \sae & $\Delta$ & \textit{p-val} \\
\midrule
$r_{\text{eff}}$ & 1.88\std{0.39} & 1.72\std{0.37} & -8.7\% & $10^{-6}$ \\
$\sigma_1/\sigma_2$ & 24.0\std{16.4} & 28.2\std{18.7} & +17.7\% & $10^{-3}$ \\
$\cos(G_t, G_{t-1})$ & 0.50\std{0.15} & 0.80\std{0.09} & +60.1\% & $10^{-36}$ \\
$\mathcal{L}$ & 0.82\std{0.45} & 1.08\std{0.38} & +31.0\% & $10^{-20}$ \\
\bottomrule
\end{tabular}
\label{tab:spectral}
\end{table}

\begin{figure}[h]
    \centering
    \includegraphics[width=\textwidth]{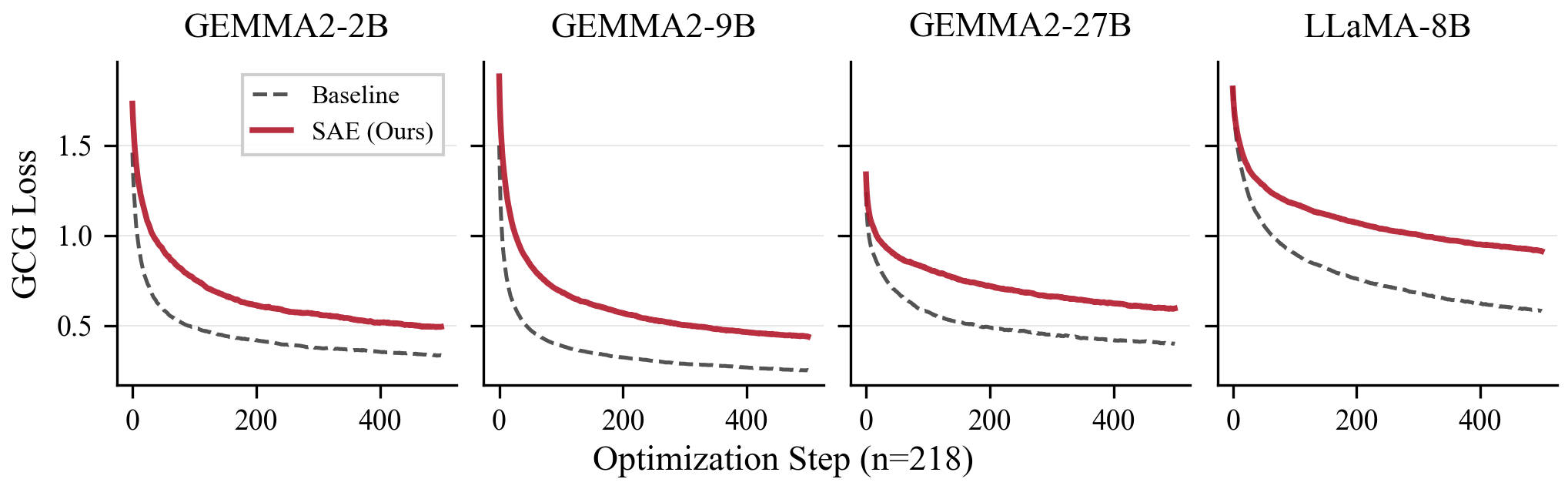}
    \caption{\gcg optimization loss curves for baseline versus \sae-intervened models. Higher final loss indicates less effective adversarial suffix generation. \sae integration increases optimization difficulty by 47--74\% across model families ($n=218$ attack runs).}
    \label{fig:gcg_loss}
\end{figure}

\looseness=-1 Across both LLaMA-8B and Gemma-9B, \sae insertion substantially alters optimization dynamics. Gradients become significantly more correlated between successive optimization steps: the cosine similarity between gradients increases from $0.50 \pm 0.15$ to $0.80\pm 0.09$ for LLaMA-8B ($p=10^{-36}$). At the same time, effective rank decreases (1.88 $\rightarrow$ 1.72) while the spectral gap $\sigma_1/\sigma_2$ increases (24.0 $\rightarrow$ 28.2), indicating that gradient signal becomes concentrated into fewer dominant directions. These changes coincide with a substantial increase in final attack loss (0.82 $\rightarrow$ 1.08, $p=10^{-20}$) and visibly slower convergence during optimization (Fig.~\ref{fig:gcg_loss}). Similar patterns are observed for Gemma-9B (Table~\ref{tab:spectral_complete}), suggesting that these effects are consistent across model families.

Together, these results suggest that \sae routing imposes a representational bottleneck on adversarial optimization. Attacks converge on a shared sparse feature subspace, but the \sae projection compresses gradient signal and increases optimization stagnation, making it harder for gradient-based methods to discover effective adversarial suffixes.
\section{Discussion}

Our results demonstrate that inserting pretrained \sae into transformer residual streams systematically improves adversarial robustness. Across model families, \sae-augmented models exhibit lower jailbreak success rates and reduced cross-model transfer under strong white-box attacks. These effects persist despite full gradient access during optimization, indicating that robustness does not arise from gradient masking, but from changes to the underlying representation space.

\xhdr{Representation-level defenses} Unlike prior defenses that operate at the input or output level (\eg paraphrasing, smoothing, or filtering), \sae insertion modifies intermediate activations. Even without retraining model weights, this lightweight intervention significantly reduces transferability of adversarial suffixes, suggesting that sparse projection disrupts the structure of transferable adversarial representations.

\xhdr{Role of sparsity and depth} Robustness depends systematically on \sae configuration. Increasing sparsity (lower $L_0$) yields a clear dose–response effect, with stronger compression leading to lower transfer ASR. Insertion depth introduces a defense–utility tradeoff: early layers provide the strongest robustness but may affect clean behavior, intermediate layers offer a balanced regime, and late layers are largely ineffective. This pattern is consistent across architectures (Gemma, LLaMA, and Qwen), indicating that it reflects general properties of transformer representations.

\looseness=-1\xhdr{Mechanistic interpretation} Feature-level analysis shows that adversarial suffixes concentrate in a shared sparse feature subspace, explaining their cross-model transferability. \sae insertion disrupts this structure. Complementary gradient analysis shows increased inter-step correlation, reduced effective rank, and larger spectral gaps, consistent with a contraction of the effective optimization subspace. Together, these results support a representational bottleneck hypothesis: sparse projection constrains the geometry available to adversarial optimization.

\xhdr{Optimization budget ablation} To assess whether extended optimization can overcome the \sae defense, we compare GCG suffixes optimized for 500 and 1500 steps on the same 50 prompts across Gemma-2 9B and LLaMA-3.1-8B-Instruct. While \sae-optimized suffixes gain substantially from additional steps—particularly in reverse transfer, where \sae$\rightarrow$Base ASR rises from 32.0\% to 38.0\% for Gemma and from 24.0\% to 64.0\% for LLaMA—the \sae defense against base-optimized suffixes remains stable or even strengthens (Base$\rightarrow$\sae: 10.0\%→4.0\% for Gemma, 34.0\%$\rightarrow$34.0\% for LLaMA), indicating that the representational bottleneck imposed by sparse routing is not easily bypassed by longer optimization alone.

\xhdr{Black-box evaluation} In black-box settings, robustness improvements depend on the evaluation protocol. \sae integration reduces ASR under classifier-based evaluation (HarmBench), while effects under other detectors are smaller. This highlights that robustness conclusions depend on how harmful behavior is defined and measured.
\section{Conclusion}
\label{sec:conclusion}
\looseness=-1 We study the robustness implications of inserting pretrained Sparse Autoencoders into transformer residual streams at inference time. Across multiple model families and scales, \sae routing consistently reduces jailbreak success rates under strong white-box attacks and decreases cross-model transferability of adversarial suffixes. Mechanistic analyses suggest that these effects arise from a representational bottleneck: sparse projection alters the geometry of adversarial optimization and concentrates gradient signal into fewer directions. These findings indicate that tools originally developed for mechanistic interpretability can also function as lightweight robustness interventions. More broadly, our results suggest that representation-level operators provide a promising direction for improving adversarial robustness in large language models without requiring retraining or gradient blocking.

\section*{Acknowledgements}
We would like to thank all the anonymous reviewers of ACL for their valuable feedback. A.S. is supported by the Fellowship in AI Research from LaCross Institute for Ethical AI in Business. C.A. is supported, in part, by grants from Capital One, LaCross Institute for Ethical AI in Business, the UVA Environmental Institute, OpenAI Researcher Program, Thinking Machine's Tinker Research Grant, and Cohere. The views expressed are those of the authors and do not reflect the official policy or the position of the funding agencies.

\bibliographystyle{unsrtnat}
\bibliography{arxiv/custom_arvix}

\appendix

\section{Experimental Setup}
\label{app:setup}

\begin{table}[h]
\centering
\caption{SAE configurations for each base model, including layer placement, model dimension, and dictionary width.}
\label{tab:sae2}
\begin{tabular}{llccc}
\toprule
\textbf{Model} & \textbf{SAE Release} & \textbf{Layer} & \textbf{$d_{\text{model}}$} & \textbf{Width} \\
\midrule
Gemma-2-2B  & gemma-scope-2b-pt-res & 12 & 2304 & 16K \\
Gemma-2-9B  & gemma-scope-9b-pt-res & 19 & 3584 & 16K \\
Gemma-2-27B & gemma-scope-27b-pt-res & 34 & 4608 & 131K \\
Llama-3.1-8B  & Llama-3.1-8B-Inst-SAE & 19 & 4096 & 65K \\
Llama-3.3-70B & Llama-3.3-70B-Inst-SAE & 50 & 8192 & 65K \\
Mistral-7B & Mistral-7B-Res-SAEs & 16 & 4096 & 65K \\
\bottomrule
\end{tabular}%
\end{table}

\begin{table*}[h]
\centering
\caption{SAE configurations used in parametric ablations (Section 5, RQ3). Layer-placement ablations vary insertion depth at fixed sparsity; sparsity ablations vary $L_0$ (Gemma) or TopK $k$ (LLaMA) at a fixed layer. For LLaMA-3.1-8B ablations, we use the \texttt{andyrdt} SAE suite, which provides multiple matched-training layer and sparsity variants not available in the canonical release.}
\label{tab:sae_ablations}
\resizebox{\textwidth}{!}{%
\begin{tabular}{lllcc}
\toprule
\textbf{Ablation} & \textbf{Model} & \textbf{SAE Release} & \textbf{Layers / Sparsity} & \textbf{Width} \\
\midrule
\multirow{3}{*}{Layer placement} 
  & Gemma-2-9B    & \texttt{gemma-scope-9b-pt-res}                  & \{5, 10, 20, 30, 35\}             & 16K  \\
  & Llama-3.1-8B  & \texttt{andyrdt/saes-llama-3.1-8b-instruct}     & \{3, 7, 11, 15, 19, 23, 27\}      & 131K \\
  & Qwen2.5-7B    & \texttt{andyrdt/saes-qwen2.5-7b-instruct}       & \{3, 7, 11, 15, 19, 23, 27\}      & 131K \\
\midrule
\multirow{2}{*}{Sparsity} 
  & Gemma-2-9B    & \texttt{gemma-scope-9b-pt-res}                  & $L_0 \in$ \{11, 36, 68, 138, 310\}& 16K  \\
  & Llama-3.1-8B  & \texttt{andyrdt/saes-llama-3.1-8b-instruct}     & $k \in$ \{32, 64, 128, 256\}      & 131K \\
\bottomrule
\end{tabular}%
}
\end{table*}

\begin{table}[h]
\centering
\caption{Refusal detection pattern categories. Full regex patterns available in supplementary code.}
\label{tab:refusal_patterns}
\begin{tabular}{ll}
\toprule
\textbf{Category} & \textbf{Example Patterns} \\
\midrule
Inability & \texttt{i can't/cannot help/assist/answer} \\
Permission & \texttt{i am not allowed/permitted/authorized} \\
Discomfort & \texttt{i don't feel comfortable/ok} \\
Policy & \texttt{my rules/guidelines prevent/prohibit me} \\
Decline & \texttt{i must/have to decline/refuse} \\
Apology & \texttt{(i) apologize/sorry, but...} \\
Appropriateness & \texttt{it's not appropriate/illegal} \\
Request rejection & \texttt{this request is unsafe/inappropriate} \\
\bottomrule
\end{tabular}%
\end{table}

\section{Additional ASR Results}
\label{app:asr}

\begin{table}[h]
\centering
\small 
\caption{Complete attack success rate (ASR) results for GCG attacks across all models and configurations. 
Base = baseline model ASR; SAE = SAE-augmented model ASR; $\Delta$ = absolute change (percentage points); 
Rel. = relative change (\%). Negative $\Delta$ indicates SAE reduced ASR; positive indicates SAE increased ASR.}
\label{tab:asr_full_appendix}
\begin{tabular}{llcccc}
\toprule
\textbf{Model} & \textbf{Config} & \textbf{Base ASR (\%)} & \textbf{SAE ASR (\%)} & \textbf{$\Delta$ (pp)} & \textbf{Rel. (\%)}  \\
\midrule
\multirow{3}{*}{Gemma 2B}
 & PROMPT & 13.8\std{2.3} & 8.3\std{1.9} & $-$5.5 & $-$40.0 \\
 & BASE & 53.2\std{3.4} & 7.1\std{1.8} & $-$46.1 & $-$86.6  \\
 & SAE & 9.6\std{2.0} & 23.9\std{2.9} & +14.2 & +147.6  \\
\midrule
\multirow{3}{*}{Gemma 9B}
 & PROMPT & 29.6\std{3.1} & 1.8\std{0.9} & $-$27.8 & $-$93.8   \\
 & BASE & 77.3\std{2.9} & 9.2\std{2.0} & $-$68.1 & $-$88.1  \\
 & SAE & 31.2\std{3.1} & 38.1\std{3.3} & +6.9 & +22.1  \\
\midrule
\multirow{3}{*}{Gemma 27B}
 & PROMPT & 52.8\std{3.4} & 33.5\std{3.2} & $-$19.3 & $-$36.5  \\
 & BASE & 74.8\std{3.0} & 38.5\std{3.3} & $-$36.2 & $-$48.5  \\
 & SAE & 27.1\std{3.0} & 41.5\std{3.4} & +14.4 & +53.2  \\
\midrule
\multirow{3}{*}{LLaMA 8B}
 & PROMPT & 18.8\std{2.7} & 26.1\std{3.0} & +7.3 & +39.0  \\
 & BASE & 49.5\std{3.4} & 22.5\std{2.8} & $-$27.1 & $-$54.6 \\
 & SAE & 26.1\std{3.0} & 14.2\std{2.4} & $-$11.9 & $-$45.6  \\
\midrule
\multirow{3}{*}{LLaMA 70B}
 & PROMPT & 21.6\std{2.8} & 10.1\std{2.0} & $-$11.5 & $-$53.2 \\
 & BASE & 43.1\std{4.1} & 18.1\std{3.2} & $-$25.0 & $-$58.1  \\
 & SAE & 42.7\std{3.4} & 5.5\std{1.6} & $-$37.2 & $-$87.1  \\
\midrule
\multirow{3}{*}{Mistral 7B}
 & PROMPT & 48.6\std{3.4} & 16.1\std{2.5} & $-$32.6 & $-$67.0  \\
 & BASE & 53.2\std{3.4} & 9.6\std{2.0} & $-$43.6 & $-$81.9  \\
 & SAE & 23.9\std{2.9} & 16.1\std{2.5} & $-$7.8 & $-$32.7  \\
\midrule
\multirow{3}{*}{\textbf{Overall}}
 & PROMPT & 30.9\std{1.3} & 16.0\std{1.0} & $-$14.9 & $-$48.2  \\
 & BASE & 59.4\std{1.4} & 17.5\std{1.1} & $-$41.9 & $-$70.5 \\
 & SAE & 26.8\std{1.2} & 23.2\std{1.2} & $-$3.6 & $-$13.4 \\
\midrule
\multicolumn{2}{l}{\textbf{All Conditions}} & 38.6\std{0.8} & 18.9\std{0.6} & $-$19.7 & $-$51.0  \\
\bottomrule
\end{tabular}
\end{table}

\begin{table*}[!h]
\centering
\small 
\caption{Complete ASR for BEAST attacks across all models and configurations. 
Base = baseline model ASR; SAE = SAE-augmented model ASR; $\Delta$ = absolute change (percentage points); 
Rel. = relative change (\%). Negative $\Delta$ indicates SAE reduced ASR; positive indicates SAE increased ASR.}
\label{tab:asr_full_appendix_beast}
\begin{tabular}{llcccc}
\toprule
\textbf{Model} & \textbf{Config} & \textbf{Base ASR (\%)} & \textbf{SAE ASR (\%)} & \textbf{$\Delta$ (pp)} & \textbf{Rel. (\%)} \\
\midrule
\multirow{3}{*}{Gemma 2B}
 & PROMPT & 13.8 & 8.3 & $-$5.5 & $-$39.9  \\
 & BASE & 13.8 & 4.5 & $-$9.3 & $-$67.4  \\
 & SAE & 20.7 & 6.6 & $-$14.1 & $-$68.1  \\
\midrule
\multirow{3}{*}{Gemma 9B}
 & PROMPT & 29.6 & 1.8 & $-$27.8 & $-$93.9  \\
 & BASE & 16.2 & 0.0 & $-$16.2 & $-$100.0  \\
 & SAE & 9.4 & 0.0 & $-$9.4 & $-$100.0  \\
\midrule
\multirow{3}{*}{Gemma 27B}
 & PROMPT & 52.8 & 33.5 & $-$19.3 & $-$36.6  \\
 & BASE & 31.2 & 19.0 & $-$12.2 & $-$39.1  \\
 & SAE & 19.2 & 16.0 & $-$3.2 & $-$16.7  \\
\midrule
\multirow{3}{*}{LLaMA 8B}
 & PROMPT & 18.8 & 26.1 & +7.3 & +38.8  \\
 & BASE & 12.4 & 12.8 & +0.4 & +3.2 \\
 & SAE & 13.8 & 17.0 & +3.2 & +23.2 \\
\midrule
\multirow{3}{*}{LLaMA 70B}
 & PROMPT & 21.6 & 10.1 & $-$11.5 & $-$53.2 \\
 & BASE & 22.5 & 9.6 & $-$12.9 & $-$57.3  \\
 & SAE & 22.5 & 10.6 & $-$11.9 & $-$52.9  \\
\midrule
\multirow{3}{*}{Mistral 7B}
 & PROMPT & 48.6 & 16.1 & $-$32.5 & $-$66.9  \\
 & BASE & 26.0 & 10.7 & $-$15.3 & $-$58.8  \\
 & SAE & 19.8 & 8.8 & $-$11.0 & $-$55.6  \\
\midrule
\multirow{3}{*}{\textbf{Overall}}
 & PROMPT & 30.9 & 16.0 & $-$14.9 & $-$48.2  \\
 & BASE & 20.4 & 9.4 & $-$11.0 & $-$53.9  \\
 & SAE & 17.6 & 9.8 & $-$7.8 & $-$44.3  \\
\midrule
\multicolumn{2}{l}{\textbf{All Conditions}} & 23.0 & 11.7 & $-$11.3 & $-$49.1  \\
\bottomrule
\end{tabular}
\end{table*}

\xhdr{Black-box results} In the black-box setting, SAE integration exhibits detector-dependent effects. As summarized in Table~\ref{tab:blackbox_asr_pm}, SAE reduces ASR under the HarmBench classifier, with mean ASR decreasing from $0.232 \rightarrow0.160$ and median ASR from $0.205\rightarrow0.116$. Under WildGuard, reductions are more modest (mean $0.448 \rightarrow 0.414$, median $0.440 \rightarrow 0.405$) and fall within overlapping confidence intervals. Under the refusal-based heuristic, ASR remains extremely high (mean $\approx 96$--$97\%$, median $\approx 99\%$) with no meaningful difference between baseline and SAE models.
\begin{table}[h]
\centering
\caption{ASR under black-box attacks across safety detectors. ``$\pm$ CI'' denotes half-width of the 95\% confidence interval.}
\begin{tabular}{llcc}
\toprule
\textbf{Detector} & \textbf{Exp} & \textbf{Mean ASR} & \textbf{Med. ASR } \\
\midrule
refusal   & B & 0.960\std{0.030} & 0.992\std{0.031} \\
refusal   & S & 0.969\std{0.023} & 0.987\std{0.013} \\
harmbench & B & 0.232\std{0.068} & 0.205\std{0.109} \\
harmbench & S & 0.160\std{0.068} & 0.116\std{0.093} \\
wildguard & B & 0.448\std{0.068} & 0.440\std{0.067} \\
wildguard & S & 0.414\std{0.073} & 0.405\std{0.081} \\
\bottomrule
\end{tabular}%
\label{tab:blackbox_asr_pm}
\end{table}
Overall, these results show that SAE show emerging robustness properties compared to their vanilla counterparts and black-box robustness depends on the operational definition of harmful behavior used by the evaluator.

\section{Cross-family validation (Qwen2.5-7B)}
\label{app:qwen}
\looseness=-1 To assess whether the observed sparsity and layer-dependent trends generalize beyond the primary model families, we evaluate Qwen2.5-7B-Instruct using a publicly available multi-layer SAE suite for a fixed checkpoint. This setup enables controlled variation of insertion depth without modifying model weights or attack procedures.
SAE insertion reduces adversarial transfer under \textsc{Base}$\rightarrow$\textsc{SAE}. At Layer 19, ASR decreases from 58.8\% to 32.4\% (Table~\ref{tab:qwen_transfer}). As in Gemma and LLaMA, transfer remains asymmetric, with \textsc{SAE}$\rightarrow$\textsc{Base} attacks exhibiting higher success rates.
Varying insertion depth reveals a strong dependence on layer position (Table~\ref{tab:qwen_layer}). Early-to-mid layers provide the strongest robustness (Layer 7: 0.0\% \textsc{Base}$\rightarrow$\textsc{SAE}), while later layers show substantially weaker effects (\eg 59.6\% at Layer 23). 
Overall, these results are consistent with the main findings: robustness is maximized when SAE routing is applied at early-to-intermediate layers and degrades for deeper insertions, suggesting that adversarial representations become more stable in later transformer layers.

\begin{table}[h] 
\centering 
\small 
\caption{Cross-configuration transfer matrix for Qwen2.5-7B-Instruct with an SAE inserted at Layer 19 (GCG, 500 optimization steps). Rows denote the evaluation model and columns denote the source of the adversarial suffix. BASE suffixes are optimized on the baseline model, while SAE suffixes are optimized on the SAE-augmented model. SAE insertion substantially reduces transfer from baseline attacks (BASE→SAE: 58.8\% → 32.4\%), while suffixes optimized on the SAE remain moderately transferable to the baseline model (SAE→BASE: 56.6\%).} 
\label{tab:qwen_transfer}
\begin{tabular}{lccc} 
\toprule 
\textbf{Evaluation Model} & \textbf{BASE Suffix} & \textbf{SAE Suffix} \\ \midrule BASE & 58.8 & 56.6 \\ SAE (L19) & 32.4 & 38.2 \\ 
\bottomrule 
\end{tabular}  
\end{table} 

\begin{table}[h] 
    \centering 
    \small 
    \caption{Effect of SAE insertion depth on adversarial transfer for Qwen2.5-7B-Instruct using a 131K-width SAE. Each row reports attack success rate (ASR) for adversarial suffixes optimized on the baseline model and evaluated on the SAE-augmented model (BASE$\rightarrow$SAE). Robustness varies strongly with layer placement: early-to-mid layers produce the strongest reduction in transfer attacks (Layer 7: 0.0\% ASR), while later layers exhibit substantially weaker defensive effects.} 
    \label{tab:qwen_layer}
    \begin{tabular}{lc} 
    \toprule 
    \textbf{Layer} & \textbf{BASE$\rightarrow$SAE ASR (\%)} \\ \midrule 3 & 31.6 \\ 7 & \textbf{0.0} \\ 11 & 11.0 \\ 15 & 16.2 \\ 19 & 32.4 \\ 23 & 59.6 \\ 27 & 53.7 \\ \bottomrule \end{tabular} 
\end{table}




\begin{table}[h]
\centering
\caption{Jaccard similarity of top-$k$ SAE features across adversarial suffix sources. 
Within-attack similarity is 3--5$\times$ higher than attack-vs-random similarity, 
indicating adversarial suffixes converge on a shared sparse feature subspace.}
\label{tab:jaccard}
\begin{tabular}{llccc}
\toprule
\textbf{Attack} & \textbf{SAE} & \textbf{Within-Atk} & \textbf{Atk-vs-Rnd} & \textbf{Ratio} \\
\midrule
GCG & Gemma-9B & 0.38\std{0.13} & 0.14\std{0.03} & 2.7 \\
 & LLaMA-8B & 0.37\std{0.17} & 0.11\std{0.06} & 3.2 \\
 & Mistral-7B & 0.31\std{0.13} & 0.06\std{0.03} & 5.0 \\
\midrule
BEAST & Gemma-9B & 0.38\std{0.11} & 0.09\std{0.04} & 4.0 \\
 & LLaMA-8B & 0.46\std{0.19} & 0.11\std{0.05} & 4.2 \\
 & Mistral-7B & 0.26\std{0.14} & 0.03\std{0.02} & 9.1 \\
\midrule
\multicolumn{2}{l}{\textbf{Combined}} & 0.36\std{0.16} & 0.09\std{0.05} & 3.9 \\
\bottomrule
\end{tabular}%
\end{table}

\begin{table}[h]
\centering
\caption{Jaccard similarity statistics across attack types and SAE models. Within-attack similarity is consistently 3--5$\times$ higher than attack-vs-random similarity.}
\label{tab:jaccard_results}
\begin{tabular}{llcccc}
\toprule
\textbf{Attack} & \textbf{SAE Model} & \textbf{$k$} & \textbf{Within-Attack} & \textbf{Within-Random} & \textbf{Attack vs Random} \\
\midrule
\multirow{3}{*}{GCG} 
 & Gemma-9B   & 135 & 0.38\std{0.13} & 0.38\std{0.12} & 0.14\std{0.03} \\
 & LLaMA-8B   & 79  & 0.37\std{0.17} & 0.28\std{0.11} & 0.11\std{0.06} \\
 & Mistral-7B & 78  & 0.31\std{0.13} & 0.32\std{0.06} & 0.06\std{0.03} \\
\midrule
\multirow{3}{*}{BEAST} 
 & Gemma-9B   & 107 & 0.38\std{0.11} & 0.40\std{0.11} & 0.09\std{0.04} \\
 & LLaMA-8B   & 83  & 0.46\std{0.19} & 0.28\std{0.11} & 0.11\std{0.05} \\
 & Mistral-7B & 59  & 0.26\std{0.14} & 0.32\std{0.07} & 0.03\std{0.02} \\
\midrule
\multicolumn{2}{l}{\textbf{GCG (Overall)}}     & -- & 0.35\std{0.15} & 0.33\std{0.11} & 0.11\std{0.05} \\
\multicolumn{2}{l}{\textbf{BEAST (Overall)}}   & -- & 0.37\std{0.17} & 0.33\std{0.11} & 0.08\std{0.05} \\
\multicolumn{2}{l}{\textbf{Combined}}          & -- & 0.36\std{0.16} & 0.33\std{0.11} & 0.09\std{0.05} \\
\bottomrule
\end{tabular}%
\end{table}


\begin{table}[t]
\centering
\small
\setlength{\tabcolsep}{3pt}
\caption{Spectral analysis of GCG gradients (Wilcoxon signed-rank test). LLaMA-8B: $n$=218; Gemma-9B: $n$=210; $\kappa$: condition number. $\text{Var}(\sigma_1)$: variance explained by first singular value.}
\begin{tabular}{clcccc}
\toprule
& \textbf{Metric} & \textbf{Base} & \textbf{SAE} & $\Delta$ & \textbf{p} \\
\midrule
\multirow{6}{*}{\rotatebox{90}{\textbf{LLaMA-8B}}} 
& $\cos(G_t, G_{t-1})$ & 0.50\std{0.15} & 0.80\std{0.09} & +60\% & $10^{-36}$ \\
& $\mathcal{L}$ & 0.82\std{0.45} & 1.08\std{0.38} & +31\% & $10^{-20}$ \\
& $r_{eff}$ & 1.88\std{0.39} & 1.72\std{0.37} & -9\% & $10^{-6}$ \\
& $\sigma_1/\sigma_2$ & 24\std{16} & 28\std{19} & +18\% & 0.005 \\
& $\kappa$ & 543\std{447} & 745\std{534} & +37\% & $10^{-8}$ \\
& $\text{Var}(\sigma_1)$ & 0.983 & 0.986 & +0\% & $10^{-4}$ \\
\midrule
\multirow{6}{*}{\rotatebox{90}{\textbf{Gemma-9B}}} 
& $\cos(G_t, G_{t-1})$ & 0.53\std{0.13} & 0.76\std{0.08} & +44\% & $10^{-33}$ \\
& $\mathcal{L}$ & 0.38\std{0.16} & 0.61\std{0.28} & +62\% & $10^{-31}$ \\
& $r_{eff}$ & 1.85\std{0.12} & 1.84\std{0.10} & -1\% & 0.644 \\
& $\sigma_1/\sigma_2$ & 8.00\std{1.00} & 8.00\std{1.00} & -2\% & 0.093 \\
& $\kappa$ & 431\std{86} & 402\std{96} & -7\% & $10^{-4}$ \\
& $\text{Var}(\sigma_1)$ & 0.975 & 0.975 & -0\% & 0.403 \\
\bottomrule
\end{tabular}
\label{tab:spectral_complete}
\end{table}



\setlength{\tabcolsep}{3pt}
\renewcommand{\arraystretch}{0.9}
\begin{table}[t]
\caption{
Attack Success Rate (ASR) with confidence intervals under black-box attacks for baseline (B) and SAE-augmented (S) models across different safety detectors. Confidence intervals for the mean are computed using Student’s \emph{t}-distribution, while confidence intervals for the median are estimated using the Hettmansperger--Sheather method.}
\centering
\small
\begin{tabular}{lcccc}
\toprule
\textbf{Detector} & \textbf{Exp} & \textbf{Mean (95\% CI)} & \textbf{Median (95\% CI)} \\
\midrule
refusal   & \base & 0.960 [0.930, 0.990] & 0.992 [0.939, 1.000] \\
refusal   & \sae      & 0.969 [0.946, 0.992] & 0.987 [0.974, 1.000] \\
harmbench & \base & 0.232 [0.165, 0.300] & 0.205 [0.133, 0.351] \\
harmbench & \sae      & 0.160 [0.092, 0.227] & 0.116 [0.051, 0.236] \\
wildguard & \base & 0.448 [0.380, 0.516] & 0.440 [0.373, 0.507] \\
wildguard & \sae      & 0.414 [0.341, 0.487] & 0.405 [0.321, 0.482] \\
\bottomrule
\end{tabular}

\label{tab:blackbox_asr}
\end{table}

\section{Comparison with other protections}
\label{app:comparing_with_other_defenses}

\looseness=-1 GCG  and BEAST exploit the smooth internal optimization landscape of LLMs to construct adversarial suffixes. Defending against such attacks therefore requires not only detecting anomalous prompts (\eg via perplexity-based filtering \cite{alon2023detectinglanguagemodelattacks}), but disrupting the feasibility of the optimization process itself.

A range of LLM defenses have been proposed \cite{yi2024jailbreakattacksdefenseslarge}. Some methods use randomized smoothing, including (Erase-and-Check \cite{kumar2025certifyingllmsafetyadversarial}, SmoothLLM \cite{robey2024smoothllmdefendinglargelanguage}, Semantic Smoothing \cite{ji2024defendinglargelanguagemodels}. Several of these methods rely on stochastic input perturbations or auxiliary LLMs for paraphrasing or response judging, increasing computational cost and introducing the need to secure the secondary model itself.

\looseness=-1 Noise-based defenses such as Smoothed Embeddings \cite{hase2025smoothedembeddingsrobustlanguage} report attack success rate (ASR) and utility trade-offs as a function of noise scale (\eg reducing GCG ASR on Vicuna from 94\% to 0\% for $\sigma \approx 0.01-0.04$. While effective, these approaches primarily evaluate aggregate ASR and utility, without analyzing how the defense reshapes optimization dynamics or cross-model transferability. Similarly, decoding-time interventions \cite{zhao2026defendinglargelanguagemodels} and adversarial training methods \cite{fu2026shortlengthadversarialtraininghelps} focus on ASR reduction, but do not characterize changes in gradient structure or the geometry of adversarial subspaces.

Our approach introduces a sparse encode–decode bottleneck into the residual stream, forcing all intermediate activations to pass through a structured sparse basis before being propagated further. This intervention preserves full gradient flow but reparameterizes the internal representation space.

\looseness=-1 As a result, adversarial suffix optimization becomes less expressive and more concentrated in a reduced set of directions. Empirically, this leads to substantially lower attack success rates (up to 5× reduction under GCG), pronounced asymmetry in cross-model transfer (suffixes optimized on baseline models transfer poorly to SAE-augmented models), and measurable changes in optimization dynamics.

In particular, we observe increased gradient alignment between successive steps, reduced effective rank of the gradient matrix, and larger spectral gaps, indicating that adversarial updates collapse into fewer dominant directions and exhibit reduced exploratory capacity. Together, these effects demonstrate that SAE integration does not merely filter outputs, but structurally reshapes the optimization landscape exploited by adversarial attacks.

\end{document}